\documentclass[10pt,twocolumn,letterpaper]{article}

\usepackage{cvpr}
\usepackage{times}
\usepackage{epsfig}
\usepackage{graphicx}
\usepackage{amsmath}
\usepackage{amssymb}

\usepackage{booktabs}

\usepackage[pagebackref=true,breaklinks=true,letterpaper=true,colorlinks,bookmarks=false]{hyperref}

\def\eg{\emph{e.g}\onedot} 
\def\ie{\emph{i.e}\onedot}

\def\wrt{w.r.t\onedot} 
\def\etal{\emph{et al}\onedot}

\newcommand{\figref}[1]{Fig\onedot~\ref{#1}}

\newcommand{\tabref}[1]{Tab\onedot~\ref{#1}}

\usepackage{pifont}
\newcommand{\cmark}{\ding{51}}

\cvprfinalcopy 

\def\cvprPaperID{****} 

\ifcvprfinal\fi
\begin{document}

\title{Panoptic-DeepLab}

\author{
Bowen Cheng$^{1,2}$, Maxwell D.~Collins$^{2}$, Yukun Zhu$^{2}$, Ting Liu$^{2}$,\\
Thomas S. Huang$^{1}$, Hartwig Adam$^{2}$, Liang-Chieh Chen$^{2}$\\
\\
{$^1$UIUC \hspace{2mm} $^2$Google Inc.}
}

\maketitle
\begin{abstract}
    We present Panoptic-DeepLab, a bottom-up and single-shot approach for panoptic segmentation. Our Panoptic-DeepLab is conceptually simple and delivers state-of-the-art results. In particular, we adopt the dual-ASPP and dual-decoder structures specific to semantic, and instance segmentation, respectively. The semantic segmentation branch is the same as the typical design of any semantic segmentation model (\eg, DeepLab), while the instance segmentation branch is class-agnostic, involving a simple instance center regression. Our single Panoptic-DeepLab sets the new state-of-art at all three Cityscapes benchmarks, reaching 84.2\% mIoU, 39.0\% AP, and 65.5\% PQ on test set, and advances results on the other challenging Mapillary Vistas.
\end{abstract}

\section{Introduction}
\label{sec:intro}

Our bottom-up {\it Panoptic-DeepLab} is conceptually simple and delivers state-of-the-art panoptic segmentation results \cite{kirillov2018panoptic}. We adopt dual-ASPP and dual-decoder modules, specific to semantic segmentation and instance segmentation, respectively. The semantic segmentation branch follows the typical design of any semantic segmentation model (\eg, DeepLab \cite{deeplabv3plus2018}), while the instance segmentation prediction involves a simple instance center regression  \cite{ballard1981generalizing,kendall2018multi}, where the model learns to predict instance centers as well as the offset from each pixel to its corresponding center.

We perform experiments on Cityscapes \cite{cordts2016cityscapes} and Mapillary Vistas \cite{neuhold2017mapillary} datasets. On Cityscapes test set, a {\it single} Panoptic-DeepLab model achieves state-of-the-art performance of 65.5\% PQ, 39.0\% AP, and 84.2\% mIoU, ranking first at  {\it all} three Cityscapes tasks when comparing with published works. On Mapillary Vistas validation set, our best {\it single} model attains 40.6\% PQ, while employing an ensemble of 6 models reaches a performance of 42.2\% PQ.
 
To summarize, our contributions are as follows.
\begin{itemize} 
    \item Panoptic-DeepLab is the {\it first} bottom-up approach that demonstrates state-of-the-art results for panoptic segmentation on Cityscapes and Mapillary Vistas.
    \item Panoptic-DeepLab is the {\it first} single model (without fine-tuning on different tasks) that simultaneously ranks first at {\it all} three Cityscapes benchmarks.
    \item Panoptic-DeepLab is {\it simple} in design, requiring only three loss functions during training, and introducing extra marginal parameters as well as additional slight computation overhead when building on top of a modern semantic segmentation model.
\end{itemize}

\section{Methods}
\label{sec:methods}

\begin{figure*}[!t]
    \centering
    \includegraphics[width=1.0\textwidth]{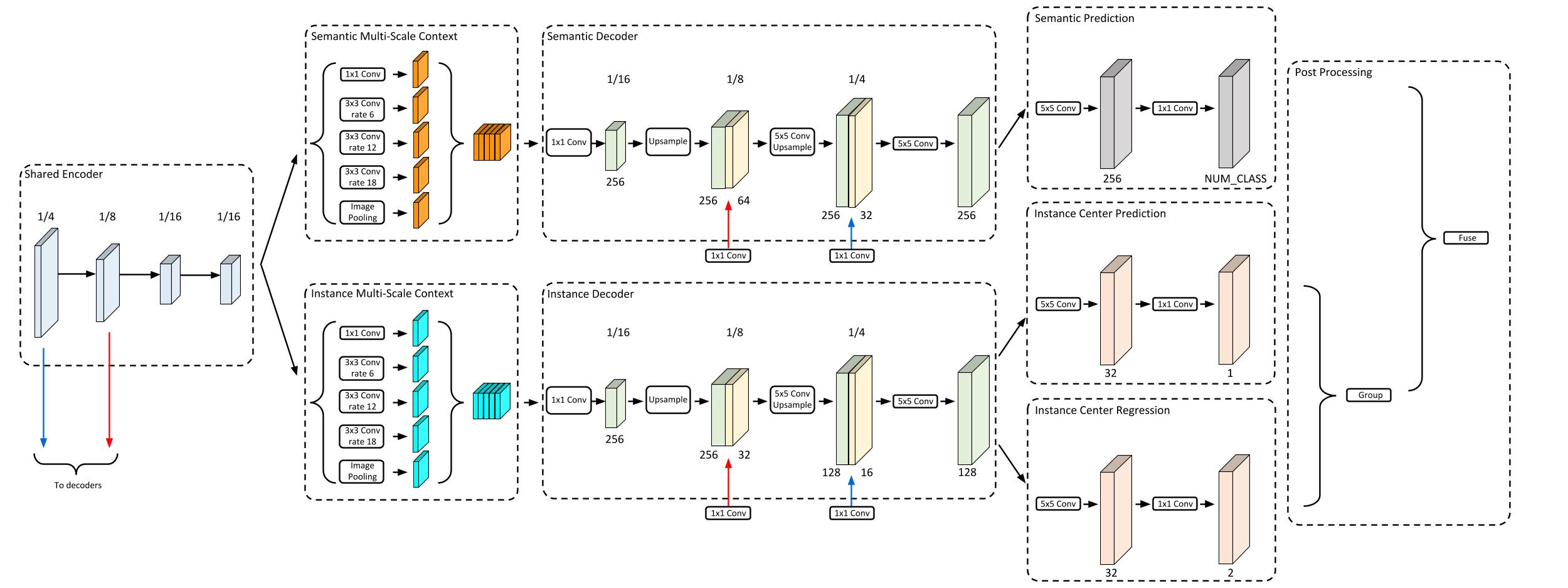}
    \caption{Our Panoptic-DeepLab adopts dual-context and dual-decoder modules for semantic segmentation and instance segmentation predictions. We apply atrous convolution in the last block of a network backbone to extract denser feature map. The Atrous Spatial Pyramid Pooling (ASPP) is employed in the context module as well as a light-weight decoder module consisting of a single convolution during each upsampling stage. The instance segmentation prediction is obtained by predicting the object centers and regressing every foreground pixel (\ie, pixels with predicted `thing' class) to their corresponding center. The predicted semantic segmentation and class-agnostic instance segmentation are then fused to generate the final panoptic segmentation result by the ``majority vote'' proposed by DeeperLab.}
    \label{fig:network_architecture}
\end{figure*}

As illustrated in \figref{fig:network_architecture}, our proposed Panoptic-DeepLab is deployed in a bottom-up single-shot manner for panoptic segmentation \cite{kirillov2018panoptic}. Panoptic-DeepLab consists of four components: (1) an encoder backbone shared for both semantic segmentation and instance segmentation, (2) decoupled ASPP modules and (3) decoupled decoder modules specific to each task, and (4) task-specific prediction heads.

{\bf Architecture:} 
The encoder backbone is adapted from an ImageNet-pretrained neural network paired with atrous convolution for extracting denser feature maps in its last block. The ASPP modules and decoder modules are separate for semantic segmentation and instance segmentation. Our light-weight decoder module gradually recovers the spatial resolution by a factor of 2; in each upsampling stage we apply only a {\it single} convolution.

{\bf Semantic segmentation:} 
We employ the typical softmax cross entropy loss for semantic segmentation.

{\bf Class-agnostic instance segmentation:} 
Motivated by Hough Voting  \cite{ballard1981generalizing,kendall2018multi}, we represent each object instance by its center of mass, encoded by a 2-D Gaussian with standard deviation of 8 pixels. For every foreground pixel (\ie, pixel whose class is a `thing'), we further predict the offset to its corresponding mass center. In particular, we adopt the Mean Squared Error (MSE) loss to minimize the distance between predicted heatmaps and 2D Gaussian-encoded groundtruth heatmaps. We use $L_1$ loss for the offset prediction, which is only activated at pixels belonging to object instances. During inference, we group predicted foreground pixels by their closest predicted mass center, forming our class-agnostic instance segmentation results.

{\bf Panoptic segmentation:} 
Given the predicted semantic segmentation and class-agnostic instance  segmentation results, we adopt a fast and parallelizable method to merge the results, following the ``majority vote'' principle proposed in DeeperLab \cite{yang2019deeperlab}. In particular, the semantic label of a predicted instance mask is inferred by the majority vote of the corresponding predicted semantic labels.
\section{Experiments}

{\bf Cityscapes \cite{cordts2016cityscapes}:} The dataset consists of 2975, 500, and 1525 traffic-related images for training, validation, and testing, respectively. It contains 8 `thing' and 11 `stuff' classes.

{\bf Mapillary Vistas \cite{neuhold2017mapillary}:} A large-scale traffic-related dataset, containing 18K, 2K, and 5K images for training, validation and testing, respectively. It contains 37 `thing' classes and 28 `stuff' classes in a variety of image resolutions, ranging from $1024\times768$ to more than $4000\times6000$

{\bf Experimental setup:} We report mean IoU, average precision (AP), and panoptic quality (PQ) to evaluate the semantic, instance, and panoptic segmentation results.

All our models are trained using TensorFlow on 32 TPUs. We adopt a similar training protocol as in~\cite{deeplabv3plus2018}. In particular, we use the `poly' learning rate policy 
with an initial learning rate of $0.001$, fine-tune the batch normalization 
parameters, perform random scale data augmentation during training, and optimize with Adam. 
On Cityscapes, our best setting is obtained by training with whole image (\ie, crop size equal to $1025\times2049$) with batch size 32. On Mapillary Vistas, we resize the images to 2177 pixels at the longest side to handle the large input variations, and randomly crop $1025\times1025$ patches during training with batch size 64. 
We set training iterations to 60K and 150K for Cityscapes and Mapillary Vistas, respectively. During evaluation, due to the sensitivity of PQ \cite{xiong2019upsnet,li2018learning,porzi2019seamless}, we re-assign to `VOID' label all `stuff' segments whose areas are smaller than a threshold. The thresholds on Cityscapes and Mapillary Vistas are 2048 and 4096, respectively. Additionally, we adopt multi-scale inference (scales equal to $\{0.5, 0.75, 1, 1.25, 1.5, 1.75, 2\}$) and left-right flipped inputs, to further improve the performance. For all the reported results, unless specified, Xception-71 \cite{deeplabv3plus2018} is employed as the backbone in Panoptic-DeepLab.

\subsection{Ablation Studies}

We conduct ablation studies on the Cityscapes validation set, as summarized in \tabref{tab:cityscapes_ablation}. Replacing the SGD momentum optimizer with the Adam optimizer yields 0.6\% PQ improvement. Instead of using the sigmoid cross entropy loss for training the heatmap (\ie, instance center prediction), it brings 1.1\% PQ improvement by applying the Mean Squared Error (MSE) loss to minimize the distance between the predicted heatmap and the 2D Gaussian-encoded groundtruth heatmap. It is more effective to adopt both dual-decoder and dual-ASPP, which gives us 0.7\% PQ improvement while maintaining similar AP and mIoU. Employing a large crop size $1025\times2049$ (instead of $513\times1025$) during training further improves the AP and mIoU by 0.6\% and 0.9\% respectively. Finally, increasing the feature channels from 128 to 256 in the semantic segmentation branch achieves our best result of 63.0\% PQ, 35.3\% AP, and 80.5\% mIoU. For reference, we train a Semantic-DeepLab under the same setting as the best Panoptic-DeepLab, showing that multi-task learning does not bring extra gain to mIoU. Note that Panoptic-DeepLab adds marginal parameters and small computation overhead over Semantic-DeepLab.

\begin{table*}[!t]
  \centering
  \scalebox{0.75}{
  \begin{tabular}{ c | c | c | c | c | c | c | c | c | c | c | c | c}
    \toprule[0.2em]
    Adam & MSE & De. x2 & ASPP x2  & L-Crop & $\text{C}_{\text{Sem}}=256$ & $\text{C}_{\text{Ins}}=256$ & Sem. Only & PQ (\%) & AP (\%) & mIoU (\%) & Params (M) & M-Adds (B) \\
    \toprule[0.2em]
    & & & & & & & & 60.3 & 32.7 & 78.2 & 41.85 & 496.84\\
    \cmark & & & & & & & & 61.0 & 34.3 & 79.4 & 41.85 & 496.84\\
    \cmark & \cmark & & & & & & & 61.8 & 33.8 & 78.6 & 41.85 & 496.84\\
    \cmark & \cmark & \cmark & & & & & & 60.8 & 32.7 & 79.0 & 41.93 & 501.88\\
    \cmark & \cmark & \cmark & \cmark & & & & & 62.5 & 33.9 & 78.7 & 43.37 & 517.17\\
    \cmark & \cmark & \cmark & \cmark & \cmark & & & & 62.7 & 34.5 & 79.6 & 43.37 & 517.17\\
    \cmark & \cmark & \cmark & \cmark & \cmark & \cmark & & & \textbf{63.0} & \textbf{35.3} & \textbf{80.5} & 46.72 & 559.15\\
    \cmark & \cmark & \cmark & \cmark & \cmark & \cmark & \cmark & & 62.1 & 35.1 & 80.3 & 46.88 & 573.86\\
    \midrule\midrule
    \cmark & & & & \cmark & \cmark & & \cmark & - & - & 80.3 & 43.60 & 518.84\\
    \bottomrule[0.1em]
  \end{tabular}
  }
  \caption{Ablation studies on Cityscapes {\it val} set. {\bf Adam}: Adam optimizer. {\bf MSE}: MSE loss for instance center. {\bf De. x2}: Dual decoder. {\bf ASPP x2}: Dual ASPP. {\bf L-Crop}: Large crop size. {\bf $\text{C}_{\text{Sem}}=256$}: 256 (instead of 128) channels in semantic segmentation branch. {\bf $\text{C}_{\text{Ins}}=256$}: 256 (instead of 128) channels in instance segmentation branch. {\bf Sem. Only}: Only semantic segmentation. M-Adds are measured \wrt a $1025\times2049$ input.}
  \vspace{-4mm}
  \label{tab:cityscapes_ablation}
\end{table*}

\subsection{Cityscapes Results}
{\bf Val set:} In \tabref{tab:cityscapes_val}, we report our Cityscapes validation set results. When using only Cityscapes {\it fine} annotations, our best Panoptic-DeepLab, with mutli-scale inputs and left-right flips, outperforms the best bottom-up approach, SSAP, by 3.0\% PQ and 1.2\% AP, and is better than the best proposal-based approach, AdaptIS, by by 2.1\% PQ, 2.2\% AP, and 2.3\% mIoU. When using extra data, our best Panoptic-DeepLab outperforms UPSNet by 5.2\% PQ, 3.5\% AP, and 3.9\% mIoU, and Seamless by 2.0\% PQ and 2.4\% mIoU. Note that we do not exploit any other data, such as COCO, Cityscapes {\it coarse} annotations, depth, or video.

{\bf Test set:} For the test set results, we additionally employ the trick proposed in \cite{deeplabv3plus2018} where we apply atrous convolution in the last two blocks within the backbone, with rate 2 and 4 respectively, during inference. We found this brings an extra 0.4\% AP and 0.2\% mIoU on {\it val} set but no improvement over PQ. We do not use this trick for the Mapillary Vistas Challenge. As shown in \tabref{tab:cityscapes_test}, our {\it single} unified Panoptic-DeepLab achieves state-of-the-art results, ranking first at {\it all} three Cityscapes tasks, when comparing with published works. Our model ranks second in the instance segmentation track when also taking into account unpublished entries.

\begin{table}[!t]
  \centering
  \scalebox{0.7}{
  \begin{tabular}{c | c | c | c | c | c | c  }
    \toprule[0.2em]
    Method & Extra Data & Flip & MS  & PQ (\%) & AP (\%) & mIoU (\%) \\
    \toprule[0.2em]
    \multicolumn{7}{c}{w/o Extra Data}\\
    \midrule
    TASCNet~\cite{li2018learning} & & & & 55.9 & - & - \\
    Panoptic FPN~\cite{kirillov2019panoptic} & & & & 58.1 & 33.0 & 75.7 \\
    UPSNet~\cite{xiong2019upsnet} & & & & 59.3 & 33.3 & 75.2 \\
    UPSNet~\cite{xiong2019upsnet} & & \cmark & \cmark & 60.1 & 33.3 & 76.8 \\
    Seamless~\cite{porzi2019seamless} & & & & 60.3 & 33.6 & 77.5 \\
    AdaptIS~\cite{sofiiuk2019adaptis} & & \cmark & & 62.0 & 36.3 & 79.2 \\
    \midrule
    DeeperLab~\cite{yang2019deeperlab} & & & & 56.5 & - & - \\
    SSAP~\cite{gao2019ssap} & & \cmark & \cmark & 61.1 & 37.3 & - \\
    \midrule\midrule
    Panoptic-DeepLab & & & & 63.0 & 35.3 & 80.5 \\
    Panoptic-DeepLab & & \cmark & & 63.4 & 36.1 & 80.9 \\
    Panoptic-DeepLab & & \cmark & \cmark & \textbf{64.1} & \textbf{38.5} & \textbf{81.5} \\
    \midrule
    \multicolumn{7}{c}{w/ Extra Data}\\
    \midrule
    TASCNet~\cite{li2018learning} & COCO & & & 59.3 & 37.6 & 78.1 \\
    TASCNet~\cite{li2018learning} & COCO & \cmark & \cmark & 60.4 & 39.1 & 78.7 \\
    UPSNet~\cite{xiong2019upsnet} & COCO & & & 60.5 & 37.8 & 77.8 \\
    UPSNet~\cite{xiong2019upsnet} & COCO & \cmark & \cmark & 61.8 & 39.0 & 79.2 \\
    Seamless~\cite{porzi2019seamless} & MV &  &  & 65.0 & - & 80.7 \\
    \midrule\midrule
    Panoptic-DeepLab & MV & & & 65.3 & 38.8 & 82.5 \\
    Panoptic-DeepLab & MV & \cmark & & 65.6 & 39.4 & 82.6 \\
    Panoptic-DeepLab & MV & \cmark & \cmark & \textbf{67.0} & \textbf{42.5} & \textbf{83.1} \\
    \bottomrule[0.1em]
  \end{tabular}
  }
  \caption{Cityscapes {\it val} set. {\bf Flip:} Adding left-right flipped inputs. {\bf MS:} Multiscale inputs. {\bf MV:} Mapillary Vistas.}
  \label{tab:cityscapes_val}
\end{table}

\begin{table}[!t]
  \centering
  \scalebox{0.7}{
  \begin{tabular}{c | c | c | c | c  }
    \toprule[0.2em]
    Method & Extra Data & PQ (\%) & AP (\%) & mIoU (\%) \\
    \toprule[0.2em]
    \multicolumn{5}{c}{Semantic Segmentation}\\
    \midrule
    Zhu~\etal~\cite{zhu2019improving} & C, V, MV & - & - & 83.5 \\
    Hyundai Mobis AD Lab & C, MV & - & - & 83.8 \\
    \midrule
    \multicolumn{5}{c}{Instance Segmentation}\\
    \midrule
    AdaptIS~\cite{sofiiuk2019adaptis} & & - & 32.5 & - \\
    UPSNet~\cite{xiong2019upsnet} & COCO & - & 33.0 & - \\
    PANet~\cite{liu2018path} & COCO & - & 36.4 & - \\
    Sogou\_MM & COCO & - & 37.2 & - \\
    iFLYTEK-CV & COCO & - & 38.0 & - \\
    NJUST & COCO & - & 38.9 & - \\
    AInnoSegmentation & COCO & - & \textbf{39.5} & - \\
    \midrule
    \multicolumn{5}{c}{Panoptic Segmentation}\\
    \midrule
    SSAP~\cite{gao2019ssap} & & 58.9 & 32.7 & - \\
    TASCNet~\cite{li2018learning} & COCO & 60.7 & - & - \\
    Seamless~\cite{porzi2019seamless} & MV & 62.6 & - & - \\
    \midrule\midrule
    Panoptic-DeepLab & & 62.3 & 34.6 & 79.4 \\
    Panoptic-DeepLab & MV & \textbf{65.5} & 39.0 & \textbf{84.2} \\
    \bottomrule[0.1em]
  \end{tabular}
  }
  \caption{Cityscapes {\it test} set. {\bf C:} Cityscapes coarse annotation. {\bf V:} Cityscapes video. {\bf MV:} Mapillary Vistas.}
  \vspace{-2mm}
  \label{tab:cityscapes_test}
\end{table}

\subsection{Mapillary Vistas Challenge}
{\bf Val set:} In \tabref{tab:mapillary_val}, we report Mapillary Vistas {\it val} set results. Our best {\it single} Panoptic-DeepLab model, with multi-scale inputs and left-right flips, outperforms the bottom-up approach, DeeperLab, by 8.3\% PQ, and the top-down approach, Seamless, by 2.6\% PQ. In \tabref{tab:mapillary_backbone}, we report our results with three families of network backbones. We  observe that na\"ive HRNet-W48 slightly under-performs Xception-71. Due to the diverse image resolutions in Mapillary Vistas, we found it important to enrich the context information as well as to keep high-resolution features. Therefore, we propose a simple modification for HRNet \cite{wang2019deep} and Auto-DeepLab \cite{liu2019auto}. For modified HRNet, called HRNet+, we keep its ImageNet-pretrained head and further attach dual-ASPP and dual-decoder modules. For modified Auto-DeepLab, called Auto-DeepLab+, we remove the stride in the original 1/32 branch (which improves PQ by 1\%). To summarize, using Xception-71 strikes the best accuracy and speed trade-off, while HRNet-W48+ achieves the best PQ of 40.6\%. Finally, our ensemble of 6 models attains a 42.2\% PQ, 18.2\% AP, and 58.7\% mIoU.

\begin{table}[!t]
  \centering
  \scalebox{0.62}{
  \begin{tabular}{c | c | c | c | c | c | c | c }
    \toprule[0.2em]
    Method & Flip & MS  & PQ (\%) & $\text{PQ}^{\text{Th}}$ (\%) & $\text{PQ}^{\text{St}}$ (\%) & AP (\%) & mIoU (\%)\\
    \toprule[0.2em]
    TASCNet~\cite{li2018learning} & & & 32.6 & 31.1 & 34.4 & 18.5 & -\\
    TASCNet~\cite{li2018learning} & \cmark & \cmark & 34.3 & \textbf{34.8} & 33.6 & \textbf{20.4} & -\\
    AdaptIS~\cite{sofiiuk2019adaptis} & \cmark & & 35.9 & 31.5 & 41.9 & - & -\\
    Seamless~\cite{porzi2019seamless} & & & 37.7 & 33.8 & 42.9 & 16.4 & 50.4\\
    \midrule
    DeeperLab~\cite{yang2019deeperlab} & & & 32.0 & - & - & - & 55.3\\
    \midrule\midrule
    Panoptic-DeepLab & & & 37.7 & 30.4 & 47.4 & 14.9 & 55.4 \\
    Panoptic-DeepLab & \cmark & & 38.0 & 30.6 & 47.9 & 15.2 & 55.8 \\
    Panoptic-DeepLab & \cmark & \cmark & \textbf{40.3} & 33.5 & \textbf{49.3} & 17.2 & \textbf{56.8} \\
    \bottomrule[0.1em]
  \end{tabular}
  }
  \caption{Mapillary Vistas {\it val} set. {\bf Flip:} Adding left-right flipped inputs. {\bf MS:} Multiscale inputs.}
  \label{tab:mapillary_val}
\end{table}

\begin{table}[!t]
  \centering
  \scalebox{0.62}{
  \begin{tabular}{c | c | c | c | c | c }
    \toprule[0.2em]
    Backbone & Params (M) & M-Adds (B) & PQ (\%) & AP (\%) & mIoU (\%) \\
    \toprule[0.2em]
    Xception-65 & 44.31 & 1054.05 & 39.2 & 16.4 & 56.9 \\
    Xception-71 & 46.73 & 1264.32 & 40.3 & 17.2 & 56.8 \\
    \midrule
    HRNet-W48~\cite{wang2019deep} & 71.66 & 2304.87 & 39.3 & 17.2 & 55.4 \\
    HRNet-W48+ & 88.87 & 2208.04 & 40.6 & 17.8 & 57.6 \\
    HRNet-W48+ (Atrous) & 88.87 & 2972.02 & 40.5 & 17.7 & 57.4\\
    HRNet-Wider+ & 60.05 & 1315.70 & 40.0 & 17.0 & 57.0 \\
    HRNet-Wider+ (Atrous) & 60.05 & 1711.69 & 39.7 & 16.8 & 56.5 \\
    \midrule
    Auto-DeepLab-L+ & 41.54 & 1493.78 & 39.3 & 15.8 & 56.9 \\
    Auto-DeepLab-XL+ & 71.98 & 2378.17 & 40.3 & 16.3 & 57.1 \\
    Auto-DeepLab-XL++ & 72.16 & 2386.81 & 40.3 & 16.9 & 57.6 \\
    \midrule\midrule
    Ensemble (top-6 models) & - & - & 42.2 & 18.2 & 58.7 \\
    \bottomrule[0.1em]
  \end{tabular}
  }
  \caption{Mapillary Vistas {\it val} set with different backbones. {\bf HRNet-W48+:} Modified HRNet-W48 with ImageNet-pretraining head kept. {\bf HRNet-W48+ (Atrous):} Additionally apply atrous convolution with rate 2 in the output stride 32 branch of HRNet. {\bf HRNet-Wider+:} A wider version of HRNet using separable convolution with large channels. The ImageNet-pretraining head is also kept. {\bf HRNet-Wider+ (Atrous):} Additionally apply atrous convolution with rate 2 in the output stride 32 branch. {\bf Auto-DeepLab-L+:} Auto-DeepLab with $F=48$ and remove the stride in the original output stride 32 path. {\bf Auto-DeepLab-XL+:} Auto-DeepLab with $F=64$ and remove the stride in the original output stride 32 path. {\bf Auto-DeepLab-XL++:} Additionally exploit low-level features from output stride 8 endpoint in the decoder module. We employ dual-ASPP and dual-decoder modules for all model variants except {\bf HRNet-W48} which follows the original design in \cite{wang2019deep}. Results are obtained with multi-scale and left-right flipped inputs. M-Adds are measured \wrt a $2177\times2177$ input.}
  \label{tab:mapillary_backbone}
\end{table}

{\small
\bibliographystyle{ieee_fullname}
\bibliography{egbib}
}

\end{document}